# USING HOPFIELD TO SOLVE RESOURCE-LEVELING PROBLEM

CAIXING LIU, JIERUI XIE[†], YUEMING HU

*College of Information Science, South China Agriculture University,*

*Guangzhou 510640, China*

*E-mail:tennic@21cn.com*

**Abstract**

Although the traditional permute matrix coming along with Hopfield is able to describe many common problems, it seems to have limitation in solving more complicated problem with more constrains, like resource leveling which is actually a NP problem. This paper tries to find a better solution for it by using neural network. In order to give the neural network description of resource leveling problem, a new description method called Augmented permute matrix is proposed by expending the ability of the traditional one. An Embedded Hybrid Model combining Hopfield model and SA are put forward to improve the optimization in essence in which Hopfield servers as State Generator for the SA. The experiment results show that Augmented permute matrix is able to completely and appropriately describe the application. The energy function and hybrid model given in this study are also highly efficient in solving resource leveling problem.

## 1. Introduction

Hopfield has advantage in solving NP-hard problems[1], such as TSP. The most common method for describing is permute matrix[2], which has been used to solve many optimization problems. The resource leveling in network planning is of great economic significance, which tries to avoid the pros and cons of resource requests in a short period of time by adjusting the starting time of each job and proves to be a NP-hard problem. But when we apply the permute matrix to the resource leveling, it shows the limitation and can not be able to give appropriate description of the problem taking into account the implications and conflicts between jobs. Therefore, we need a more capable method to describe the application before finding a Neural Network solution for it.

### *1.1 Augmented Permute Matrix*

Augmented permute matrix (APM) is introduced in this study, which is based on the traditional permute matrix. The elements in the APM can only take "1" or "0" as well. The difference and what makes it more powerful is that APM is allow to have arbitrary amount of "1". That is to say, it can have no "1", one "1" or many "1". The columns and the rows in APM are completely independent and parallel, Fig.1 shows some examples of APM. When the rule "only one '1' in each row and each column" is applied, it functions as the traditional one. The APM is more powerful in description, so it can be regard as an addition to the existing describing methods[3].

To a resource leveling problem with m jobs and limit of n days, APM is m×n order. Each row represents one job; the value 1 stands for the activation of job in certain day; then number of 1 should equal to $J_x$, the duration of the job; no. 0 is allowed to appear in any continuous 1s to make the non interrupt constraint.

---

[†] Corresponding author.

*Email address*: tennic@21cn.com (Jierui XIE)

There is no special rule for columns in the resource leveling problem. They simply represent the amount of resource that is requested in certain days. Let each element in APM have a neuron counterpart, so m×n neurons are needed to describe the resource leveling problem in Hopfield. Accordingly, a weight matrix with (m×n)^2 elements is needed.

|   | 1 | 2 | 3 | 4 | 5 |
|---|---|---|---|---|---|
|   | 1 | 0 | 0 | 0 | 0 |
| 2 | 0 | 0 | 0 | 0 | 0 |
| 3 | 1 | 1 | 1 | 0 | 0 |
| 4 | 1 | 1 | 1 | 1 | 0 |
|   | 1 | 1 | 1 | 1 | 1 |

|   | 1 | 2 | 3 | 4 | 5 |
|---|---|---|---|---|---|
|   | 1 | 0 | 1 | 1 | 1 |
| 2 | 0 | 0 | 1 | 1 | 1 |
| 3 | 0 | 0 | 1 | 1 | 1 |
| 4 | 0 | 0 | 0 | 1 | 1 |
| 5 | 0 | 0 | 0 | 0 | 1 |

Fig.1 Examples of

## 2. Design for Energy Function

We adopt discrete Hopfield. The reason for using DHNN is because it can naturally satisfy the 0/1 constrain. This advantage allows most other constrains become restricted and independent by eliminating the internal relations against getting steady. The following designs for original object and constrains are based on penalty method.

（1）**Energy Term for Object Function Constrain**

$$E_0 = \frac{M}{2}\sum_{i=1}^{n}(\sum_{x=1}^{m} Q_x \cdot v_{xi} - Q_m)^2 \quad (1)$$

where $M$ is nonnegative penalty parameter，$Q_x$ is the amount of resource that $x$ th job requests，$Q_m$ is a constant representing the average resource request, $v_{xi}$ stands for the output of neuron at row $x$ and column. Daily resource request can be calculated by multiplying neuron in the column by its correspondent resource request. If $v_{xi}=1$, it means job $x$ is active on $i$th day, $Q_x \times v_{xi}$ equals to $Q_x$ representing the actual resource occupied; otherwise, $Q_x \times v_{xi}=0$.

（2）**Energy Term for Row Constrain**
As stated above, each row in APM represents one job with elements can only be 1 or 0.Where 1 stands for active state and 0 stands for non-active state of a certain job on certain day. Furthermore the number of 1 should be restricted to duration ,so the energy term can be defined as follows：

$$E_1 = \frac{A}{2}\sum_{x=1}^{m}(\sum_{i=1}^{n} v_{xi} - J_x)^2 \quad （2）$$

Where A is nonnegative penalty parameter.

（3）**Energy Term for Non-Interrupt Constrain**
To distinguish different continuous state of a sequence of Non-Interrupted 1 sequence (such as In the middle、Begin from 1st day、End on the last day and Through the whole limit of time) and unexpected fake continuous state, "Virtual Column/Row "(VCR) is proposed. VCR is to adding new columns or rows to the original APM. The additive column or row is set all zeros and just to facilitate the design not representing any actual functional neuron or not connecting with other neurons. It also has no impact on the weight matrix. For the above feature, it takes the name of "Virtual". A column is appended to the APM which finally helps to achieve the construction of Non-Interrupt Constrain energy term. The energy term can be defined as follows：

$$E_2 = \frac{F}{2}\sum_{x=1}^{m}\sum_{i=1}^{n}(v_{xi} - v_{x,\text{mod}(i,n)+1})^2 \quad (3)$$

Where F is nonnegative penalty parameter. By means of the character of DHNN,Calculate the square sum of the result each neuron subtracting its close behind neighbor and take it as the penalty for the constrain. It is noted that when refer to the last neuron concept of "loop" is applied by using "mod" operator.

### (4) Energy Term for Front Relation Constrain

In term of comparing the position of two jobs, determining they starting time is first and foremost, that is, to find out the position of the first 1 element in each row. To achieve this goal, new methods derived from binary coding are introduced which are named "forward coding" and "backward coding". Auxiliary code is a permutation of 2's power. The permutation sorted ascendingly, such as [21 22…2n], is "forward coding". Contrarily, Permutation like [2n 2n-1…21] is called "backward coding".

Let the number of permutation equal to that of neurons and makes a map between the two, operations (such as multiply) can be take as necessary. The energy term can be defined as follows:

$$E_3 = J \cdot \sum_{x}^{m}\sum_{y}^{m}(pos_y - pos_x) \cdot \text{sgn}(pos_y - pos_x) \cdot FrontT_{xy} \quad (4)$$

Where *J* is nonnegative penalty parameter. *FronT* is a table reflecting the front relation of jobs. It serves as an auxiliary: suppose there are m jobs, then the *FrontT* is square matrix with m×m order.

If job *y* is in the front of job *x* in the network planning, $FronT_{xy}$ (the position x row y column) is set 1, otherwise, set 0. Specially, *FrontT* is asymmetric, namely $Front_{xy}$ and $Front_{yx}$ mean difference.

In addition, where $Pos_x$ represents the position of the first 1 element,

$$\text{sgn}(x) = \begin{cases} 0, x \prec 0 \\ 1, x \geq 0 \end{cases}$$

It should be noted that Front Relation Constrain term will adding to the total energy in which decide whether accept *Meropolis* rule, but it will not impact on the network structure (no according weights and biases).

Combining all the terms given above, the complete energy function of resource leveling can be form as the following:

$$E = \frac{M}{2}\sum_{i=1}^{n}(\sum_{x=1}^{m}Q_x \cdot v_{xi} - Q_m)^2 + \frac{A}{2}\sum_{x=1}^{m}(\sum_{i=1}^{n}v_{xi} - J_x)^2 + \frac{F}{2}\sum_{x=1}^{m}\sum_{i=1}^{n}(v_{xi} - v_{x,\text{mod}(i,n)+1})^2$$
$$+ J \cdot \sum_{x}^{m}\sum_{y}^{m}(pos_y - pos_x) \cdot \text{sgn}(pos_y - pos_x) \cdot FrontT_{xy} \quad (5)$$

The correspondent weights and biases can be given as follows:

$$\begin{cases} T_{xi,yj} = -M \cdot Q_x \cdot Q_y \cdot \delta_{ij} - A \cdot \delta_{xy} - F \cdot \delta_{xy} \cdot \delta_{ij} + 2F \cdot \delta_{xy} \cdot \delta_{j,\text{mod}(i,n)+1} \\ I_{xi} = M \cdot Q_x \cdot \overline{Q_m} + A \cdot J_x \end{cases}$$

### 3. DHNN-SA Embedded Hybrid Model

*3.1 Basic Theory*

DHNN-SA is characterized by using the Hopfield to serve as the state generator for SA. The concept of Embedded Hybrid Model in this study is totally different from the reference [4] means. The basic theory is as follows: during the process of optimization, the neurons update their states under the force of weights and biases matrix in iteration, and provide a new state for the SA in the solution space; energy of the system is calculated; SA takes Metropolis rule to decide whether it accepts the new state or not, after that the temperature is adjusted according to some descending rules. If the stop criterion has been reached, then stop the network, otherwise, Hopfield goes on with iteration from the new state. Therefore, from the view of space, DHNN and SA acts independently on one hand, and on the other, they join each other in some intersections. Property of joining but not complete including is the distinct character of the new hybrid model.

Even in the hybrid model, DHNN has somehow independent and follow its own iteration. Moreover the DHNN here is not restricted to go down in an ever-decreasing energy track. A fluctuant one will be accepted. Actually, if the energy descent constantly during the iteration, SA will not work for the condition that the energy is reduced all the time and the bad condition that energy increased will never occurs. That is, the bad state will not be reached with zero probability in term of Metropolis. In this case, SA serves as the upper bound of the iterations. DHNN-SA is almost equal to single DHNN.

Only under the circumstance where single DHNN's energy fluctuates in the operation, DHNN-SA shows its power. To the single DHNN, this condition generally indicates an unsteady energy function, but now it should also accepted new state under the SA mechanism, and takes the possibility to jump out of the local minimum. Its track of energy is expected to be fluctuating but decreasing from the whole.

The Embedded Hybrid Model shows great advantage by using Hopfield to make up SA's shortcoming in slow iteration and utilizing SA to bring Hopfield out from local minimum. It provides a new, effective, and optional approach for complex problem such as Resource Leveling. Even the theoretical steady condition of Hopfield can not meet, it helps to get a satisfying solution.

*3.2 Algorithm Description*

Next, program can be derived from the above describing energy function and runs to get the optimizing results. The following program is bases on penalty method, taking SA as outer loop and discrete-time Hopfield as inner loop. As can be seen, DHNN-SA is the key component of the program. Algorithm Description is in Fig.2.

```
DHNN-SA()
{
   stop=0;
   T=T_0;
   GetInitialVxi();
   E_old=CaculateEnergy();
     while(~stop) {
         for (Step=1:L) {
             newVxi=UpdateNet();
             E_new = CaculateEnergy();
             △E=E_new-E_old;
             if (△E<0) {
                 oldVxi=newVxi;
                 E_old = E_new;}
             else {
                 if( exp(-△E /(K*T))<= rand())
                     Continue;
                 Else {
                     oldVxi=newVxi;
                     E_old = E_new;}
             }
         }
       T = DecreasingRule(T);
       If   (T<=E_stop) stop=1;
   }
}
```

**Fig.2 Egergy-Time curve of Exp.1**

能量图

## 4. Experiments and Results Analysis

The following experiment based on the instance from reference [5]. The data are from practical construction project that has 9 jobs and the time is limited to 14 days. The key parameters that are used in the experiments include DHNN parameters and SA parameter. DHNN parameters are given as M=1，A=300，F=200，Z=200，J=200; SA parameters are given as：L=1000 T0=100, a=0.90. where L is the length of Makoval chain，that is the steps for inner loop; T0 is the initial temperature; a is the rate for controlling the descending speed of temperature. The decreasing process is controlled by linear function like T'=a×T. Table 1 shows the results of five experiments running from different initial state. Since the tracks of energy with time elapsing are similar in the experiments, only the NO.1's is presented in Fig. 3.

Table 1 the results from 5 experiments of DHNN-SA

| No | Itera-Stpes | Itera- Time | Mini-Energy | Variance |
|----|-------------|-------------|-------------|----------|
| 1  | 3200        | 30.92       | 1825.83     | 3.69     |
| 2  | 3200        | 32.15       | 1839.85     | 9.26     |
| 3  | 1900        | 33.87       | 1838.81     | 12.83    |
| 4  | 2700        | 14.10       | 1825.81     | 3.69     |
| 5  | 1600        | 15.70       | 1851.83     | 10.40    |
|    | Avg-Steps   | Avg- Time   | Avg-Energy  | Avg-Variance |
|    | 2520        | 25.34       | 1836.42     | 7.97     |

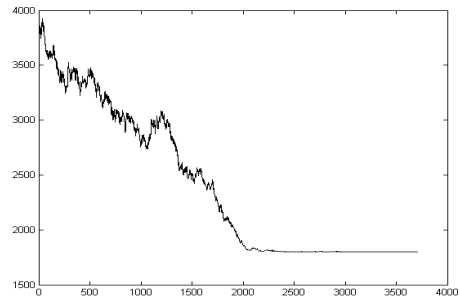

Fig.3 Egergy-Time curve of Test

In order to shows the advantage of DHNN-SA in solving resource leveling, a comparison between DHNN-SA and PM2002（Microsoft Project 2002）and P3(Primavera Project Planner)are presented below also. Because PM2002 and P3 represent professional project management software, so the results seem to be more typical. Table 2 contains the best optimizing result from the three.

Table 2 Best results from DHNN-SA、PM2002 and P3

| Meth \ Day | 1 | 2 | 3 | 4 | 5 | 6 | 7 | 8 | 9 | 10 | 11 | 12 | 13 | 14 | Variance |
|---|---|---|---|---|---|---|---|---|---|---|---|---|---|---|---|
| Original | 14 | 14 | 19 | 19 | 20 | 8 | 12 | 12 | 12 | 12 | 9 | 5 | 5 | 5 | 24.41 |
| PM2002 | 14 | 14 | 12 | 12 | 16 | 15 | 16 | 12 | 12 | 12 | 12 | 9 | 5 | 5 | 11.13 |
| P3 | 11 | 11 | 12 | 12 | 12 | 14 | 16 | 16 | 12 | 12 | 12 | 12 | 9 | 5 | 6.83 |
| DHNN-SA | 14 | 14 | 12 | 12 | 9 | 11 | 12 | 16 | 9 | 9 | 12 | 12 | 12 | 12 | 3.68 |

As can be seen from Table 1, all the five optimization results from DHNN-SA are obviously that of original network (variance less than 24.41); The analysis from Table 1 and 2 is as following: the variance of the best result that PM2002 gains is 11.13, but the average level of DHNN-SA is 7.97，near 71.90% of the former; P3 can reach the result as better as 6.83, that is, 1.14 less than average level of DHNN-SA. However, there are two experiments out of the five get 3.69, twice as better as P3, namely 40% of the experiments gain high quality solutions. It should be noted that as indicated in the Fig. 1, No. 1 and 4 haven't reached the minimum of the energy in fact, which are just the best approximate solution obtained in the above experiments under the maximum iteration steps. Given the decreasing steep more slower （a＞0.9），initial temperature higher ($T_0$>100) and Markvol chain longer (L>1000), it should be able to find the higher quality solutions。

## 5. Conclusion and Future Work

A new concept of Augmented permute matrix is proposed to describe the resource leveling problem. Then an Embedded Hybrid Model combining Hopfield model and SA in which Hopfield servers as State Generator for the SA is put forward to map the whole application. The recent experiments indicate that the neural network solution is highly efficient.

Nevertheless, only the "single resource time limited" case of resource leveling is discussed in the current study. But the scheme and all the design technologies are useful and powerful in other application, and will pay attention to the multi-resource[6] and multi-goal problem in the resource leveling field in the future research.

## References

[1] Hopfield J.J. Neurons with graded response have collective computational properties like those of two-state neurons[J]. Proceedings of the National Academy of Sciences (USA),1984,(81):3088-3092

[2] D. W. Tank (HP) and J. J. Hopfield (HP).Simple 'neural' optimization networks: an A/D converter, signal decision circuit, and a linear programming circuit[J]. IEEE Transactions on Circuits and Systems, 1986,(33):533~541.

[3] Takeda M,Goodman JW. Neural Networks for computation: Number Representation and programming complexity[J]. Applied Optics, 1986,25(18)3033~3039.

[4] Metropolis N, Rosenbluth A, Rosenbluth M et al. Equation of state calculations by fast computing machines[J]. Journal of Chemical Physics, 1953(21):1087~1092.

[5] Wang Nuo. Research on Network planning and its Application[M].Beijing:China Communications Press,1999

[6] Stinson JP.. Multiple resource-constrained scheduling using branch and bound[J] .ALLE Trans,2000,(10):16-22